\documentclass[12pt]{article}

\usepackage[utf8]{inputenc}
\usepackage[english]{babel}
\usepackage[utf8]{inputenc}
\usepackage{amsmath}
\usepackage{float}
\usepackage{tcolorbox}
\usepackage{graphicx}
\usepackage[margin=.5in]{geometry}
\usepackage[colorinlistoftodos]{todonotes}
\usepackage{amsmath,amssymb,dsfont,amsthm}
\usepackage{graphicx}
\usepackage{dsfont}
\usepackage{listings}	
\usepackage{color}
\usepackage{bm}
\usepackage{wasysym}
\usepackage[colorinlistoftodos]{todonotes}
\usepackage{mathtools}
\usepackage{caption}
\usepackage{subcaption}
\usepackage{hyperref}
\usepackage{gensymb}

\usepackage{microtype}        
\usepackage{ccicons}          

\usepackage{todonotes}

\title{Detection and Mitigation of Bias in Ted Talk Ratings}
\author{Rupam Acharyya*,Shouman Das*,Ankani Chattoraj, Oishani Sengupta,Md Iftekar Tanveer}

\begin{document}

\maketitle

\begin{abstract}
  Unbiased data collection is essential to guaranteeing fairness in artificial intelligence models. Implicit bias, a form of behavioral conditioning that leads us to attribute predetermined characteristics to members of certain groups and informs the data collection process. This paper quantifies implicit bias in viewer ratings of TEDTalks, a diverse social platform assessing social and professional performance, in order to present the correlations of different kinds of bias across sensitive attributes. Although the viewer ratings of these videos should purely reflect the speaker's competence and skill, our analysis of the ratings demonstrates the presence of overwhelming and predominant implicit bias with respect to race and gender. In our paper, we present strategies to detect and mitigate bias that are critical to removing unfairness in AI.

\end{abstract}





\section{Introduction}
Machine-learning techniques are being used to evaluate human skills in areas of social performance, such as automatically grading essays~\cite{alikaniotis2016automatic,taghipour2016neural}, outcomes of video based job interviews~\cite{chen2017automated,Naim2016}, hirability~\cite{Nguyen2016}, presentation performance~\cite{Tanveer2015,Chen2017a,Tanveer2018} etc. These algorithms automatically quantify the relative skills and performances by assessing large quantities of human annotated data. Companies and organizations world-wide are increasingly using commercial products that utilize machine learning techniques to assess these areas of social interaction. However, the presence of implicit bias in society reflected in the annotators and a combination of several other unknown factors (e.g. demographics of the subjects in the datasets, demographics of the annotators) creates systematic imbalances in human datasets. Machine learning algorithms (neural networks in most cases) trained on such biased datasets automatically replicate the imbalance~\cite{o2016weapons} naturally present in the data and result in producing \emph{unfair} predictions.

Examining the impact of implicit bias in social behavior requires extensive, diverse human data that is spontaneously generated and reveals the perception of success in social performance. In this paper, we analyze ratings of TED Talk videos to quantify the amount of social bias in viewer opinions. TED Talks present a platform where speakers are given a short time to present inspiring and socially transformative ideas in an innovating and engaging way. In its mission statement, the TED organization describes itself as a ``global community, welcoming people from every discipline and culture'' and makes an explicit commitment to ``change attitudes, lives, and ultimately the world'' \cite{tedtalk}.

Since TED Talks offer a platform to speakers from diverse backgrounds trying to convince people of their professional skills and achievements, the platform lends itself to a discussion of several critical issues regarding fairness and implicit bias: How can we determine the fairness of viewer ratings of TED Talk videos? Can we detect implicit bias in the ratings dataset? Are ratings influenced by the race and gender of the speaker? Ideally, these ratings should depend on the perception of the speaker's success and communicative performance; not on the speaker's gender or ethnicity. For instance, our findings show that while a larger proportion of viewers rate white speakers in a confidently positive manner, speaker of other gender identities and ethnic backgrounds receive a greater number of mixed ratings and elicit wider differences of opinion. In addition, men and women are rated as positive or negative with more consistency, while speakers identifying with other gender identities are rated less consistently in either direction. With this assumption, we conducted computational analysis to detect and mitigate bias present in our data. We utilize a state of the art metric ``\textit{Disparate Impact}'' as in~\cite{feldman2015certifying} for measuring fairness, and three popular methods of bias correction---
\begin{enumerate}\item pre-processing~\cite{calmon2017optimized,kamiran2012data}, \item in-processing~\cite{calders2010three,kamishima2011fairness}, and \item post-processing~\cite{hardt2016equality} 
\end{enumerate}. 
We compared such predictions of the ratings with the actual ratings provided by the viewers of the TED talks and found that our model prediction performs better w.r.t. a standard fairness metric.

Our experiments show that if the traditional machine learning models are trained on a dataset without any consideration of the data bias, the model will make decision in an unwanted way which could be highly unfair to an unprivileged group of the society. In short, major contributions of the paper are as follows,

\begin{enumerate}
    \item We show that public speaking ratings can be biased depending on the race and gender of a speaker. We utilize the state-of-the-art fairness measuring metric to identify the biases present in the TED talk public speaking rating. 
    
    \item We propose a systematic procedure to detect unfairess in the TEDTalk public speaking ratings. This method can be adopted by any machine learning practitioner or data scientist at industry who are concerned with fairness in their datasets or machine learning models.
    
\end{enumerate}

\section{Related Works}
With the increased availability of huge amount of data, data-driven decision making has emerged as a fundamental practice to all sorts of industries. In recent years, data scientists and the machine learning community put conscious effort to detect and mitigate bias from data sets and respective models. Over the years, researchers have used multiple notions of fairness as the tools to get rid of bias in data that are outlined below: 
\begin{itemize}
    \item \emph{`individual fairness'}, which means that similar individuals should be treated similarly~\cite{dwork2012fairness} 
    \item \emph{`group fairness'}, which means that underprivileged groups should be treated same as privileged groups~\cite{pedreschi2009measuring,pedreshi2008discrimination}.
    \item \emph{`fairness through awareness'}, which assumes that an algorithm is fair as long as its outcome or prediction is not dependent on the use of protected or sensitive attributes in decision making~\cite{grgic2016case}.
    \item \textit{`equality of opportunity'}, mainly used in classification task which assumes that the probability of making a decision should be equal for groups with same attributes~\cite{hardt2016equality}. 
    \item \textit{`counterfactual fairness}, very close to equality of opportunity but the probability is calculated from the sample of counterfactuals~\cite{russell2017worlds, kusner2017counterfactual} which ensures that the predictor probability of a particular label should be same even if the protected attributes change to different values . 
\end{itemize}
The fairness measures mentioned above can be characterized as both, manipulation to data and implementation of a supervised classification algorithm. One can employ strategies of detecting unfairness in a machine learning algorithm, observe~\cite{zliobaite2015survey} and removing them by, 
\begin{itemize}
    \item Pre-processing: this strategy involves processing the data to detect any bias and mitigating unfairness before training any classifiers~\cite{calmon2017optimized,kamiran2012data}.
    \item In-processing: this technique adds a regularizer term in the loss function of the classifier which gives a measurement of the unfairness of the classifier~\cite{calders2010three,kamishima2011fairness}.
    \item Post-processing: this strategy manipulates predictor output which makes the classifier fair under the measurement of a specific metric~\cite{hardt2016equality}.
\end{itemize} 
For our analysis, we follow this well established paradigm 
and use an open-source toolkit AIF360~\cite{aif360-oct-2018} to detect and mitigate bias present in the data set and classifiers at all three stages: the pre-processing, the in-processing and the post-processing step.
\section{Data Collection}
\begin{figure}
\centering
\includegraphics[width=1.\columnwidth]{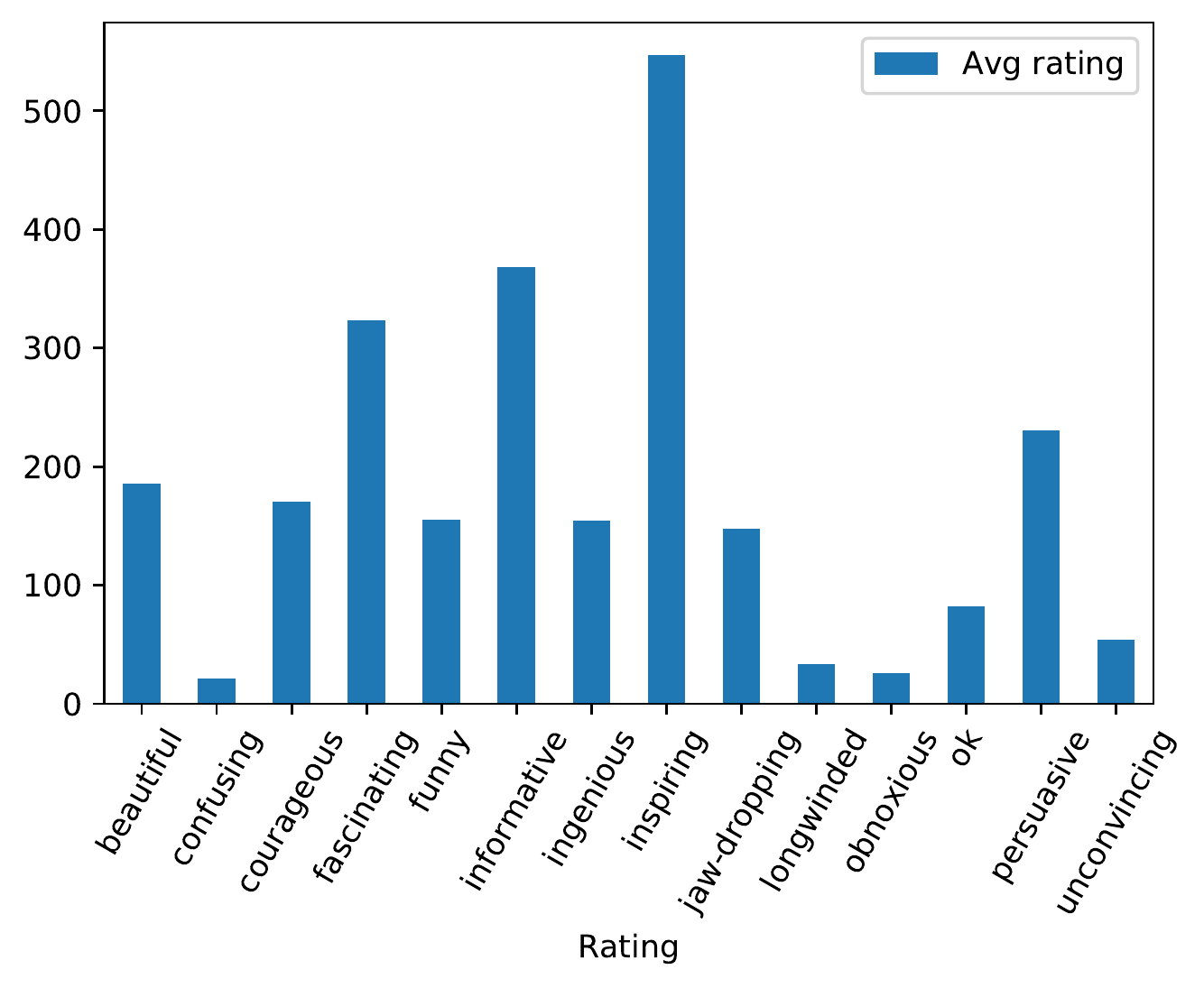} 
\caption{Average number of ratings per video in each of the fourteen categories in the TedTalk dataset. observe that the the positive ratings outnumbers the negative rating on average which inform us about the pattern of the viewer's rating. Viewer's are more inclined to give positive ratings. }
\label{fig:avg_rating}
\end{figure}

We analysed the TedTalk data collected from the \url{ted.com} website. We crawled the website and gathered information about TedTalk videos which have been published on the website for over a decade (2006-2017). These videos cover a wide range of topics, from contemporary political, social issues to modern technological advances. The speakers who delivered talks at the TedTalk platform are also from a diverse background; including but not limited to, scientists, education innovators, celebrities, environmentalists, philosophers, filmmakers etc. These videos are published on the \url{ted.com} website and are watched by millions of people around the world who can give ratings to the speakers. The rating of each talk is a collection of fourteen labels such as beautiful, courageous, fascinating etc. In this study we try to find if there is any implicit bias in the rating of the talks with respect to the race and gender of the speaker. Some properties of the full dataset is given in table \ref{tab:datasize}. Each viewer can assign three out of fourteen labels to a talk and we use the total count for each label of rating for our analysis. In figure \ref{fig:avg_rating}, average number of ratings in each of the fourteen categories is shown as a bar plot. Our preliminary observation reveals some disparities among the rating labels e.g. the label `inspiring' has significantly higher count than other labels.
\begin{table}
  \begin{center}
  \begin{tabular}{|c|c|}
    \hline
    \textbf{Property}& \textbf{Quantity}\\
    \hline
    Total number of Talks & 2,383\\
    \hline
    Average number of views per talk & 1,765,071\\
    \hline
    Total length of all talks & 564.63 Hours\\
    \hline
    Average rating labels per talk & 2,498.6\\
    \hline
\end{tabular}
\end{center}
  \caption{ TED talk Dataset Properties: Information about the TED talk videos that are used in our method of detecting unfairness}
  \label{tab:datasize}
\end{table}

\subsection{Data Annotation and Normalization}
Our raw data includes information about the topic of the talk, number of views, date of publications, the transcripts used by the speakers, count of rating labels given by the viewers etc. However, this raw data does not come with the protected features such as gender and race of the speakers.  We identified gender and race as potential sensitive attributes from our preliminary analysis. Thereby, to annotate the data for these protected features, we used amazon mechanical turk. We assigned three turkers for each talk, and annotated the race and gender of the speaker. To estimate the reliability of the annotation, we performed the Krippendorff's alpha reliability test~\cite{krippendorff2018content} on the annotations. We found a value of 93\% for the Krippendorff's alpha on our annotations. If there is a disagreement among the annotations of different turkers, we take the majority vote whenever possible or, we manually investigate to do the annotations. For gender, we used three categories: male, female, others, and for race, we used four categories, White, African American or Black, Asian, others.
In our analysis, we used the total number of views $(V)$, the transcripts of the talks $(T)$ and the rating $(Y)$ given by the viewers. For preprocessing we employ the following steps,
\begin{itemize}
    \item First we normalize the number of views using the min-max normalization technique. 
$$
V = \frac{V-V_{\min}}{V_{\max}-V_{\min}}
$$
    \item For handling the transcripts, we utilize a state-of-the-art text embedding method `doc2vec' as described in ~\cite{le2014distributed}
    \item We also scale the rating $(Y)$ of each video. Note that $Y$ consists of fourteen numbers $(y_1,\cdots, y_{14})$ which represents the number of fourteen rating categories. We scale it as, 
    $$
    Y_{\textrm{scaled}} = \frac{Y}{\sum_{i=1}^{14}y_i}
    $$
    Finally, to train our classification model, we binarize our rating lables with a threshold of the median across all talks. This means that for a given talk the rating label beautiful can be 0 or 1, courageous label can be 0 or 1 and so on for all fourteen possible lables per talk. Here 0 indicates the label does not hold and 1 indicates that the label holds true for the respective talk.
\end{itemize}

\begin{table}
  \begin{center}
  \begin{tabular}{|c|c|c|c|}
    \hline
    \textbf{Gender}& \textbf{Count} & \textbf{Race}& \textbf{Count}\\
    \hline
    Female & 768 & White & 1901\\
    \hline
    Male & 1596 & Asian &210\\
    \hline
    Other & 19 & Black & 169\\
    \hline
    \ & \ & Other & 103\\
    \hline
\end{tabular}
\end{center}
  \caption{ Total count of protected attributes (race and gender) in the TedTalk dataset. One preliminary obeservation is white male outnumbers all other groups by a huge margin, also third gender speakers are underrepresented in the data set.}
  \label{tab:protected_attrinutes}
\end{table}

\section{Observation in Data}
Before we train our machine learning models for predicting TedTalk rating, we performed several exploratory analysis on the data set. First, we counted the total number of different gender and races of the speakers, see Table \ref{tab:protected_attrinutes}. We notice that there is a significant imbalance in the gender and race count. The count of `white male' speakers outnumbers all the other groups combined.  Also, the number of speakers from the third gender community is very small which shows an noteworthy imbalance in the dataset. \\ 
\begin{figure*}
\centering
\includegraphics[width=1.\linewidth]{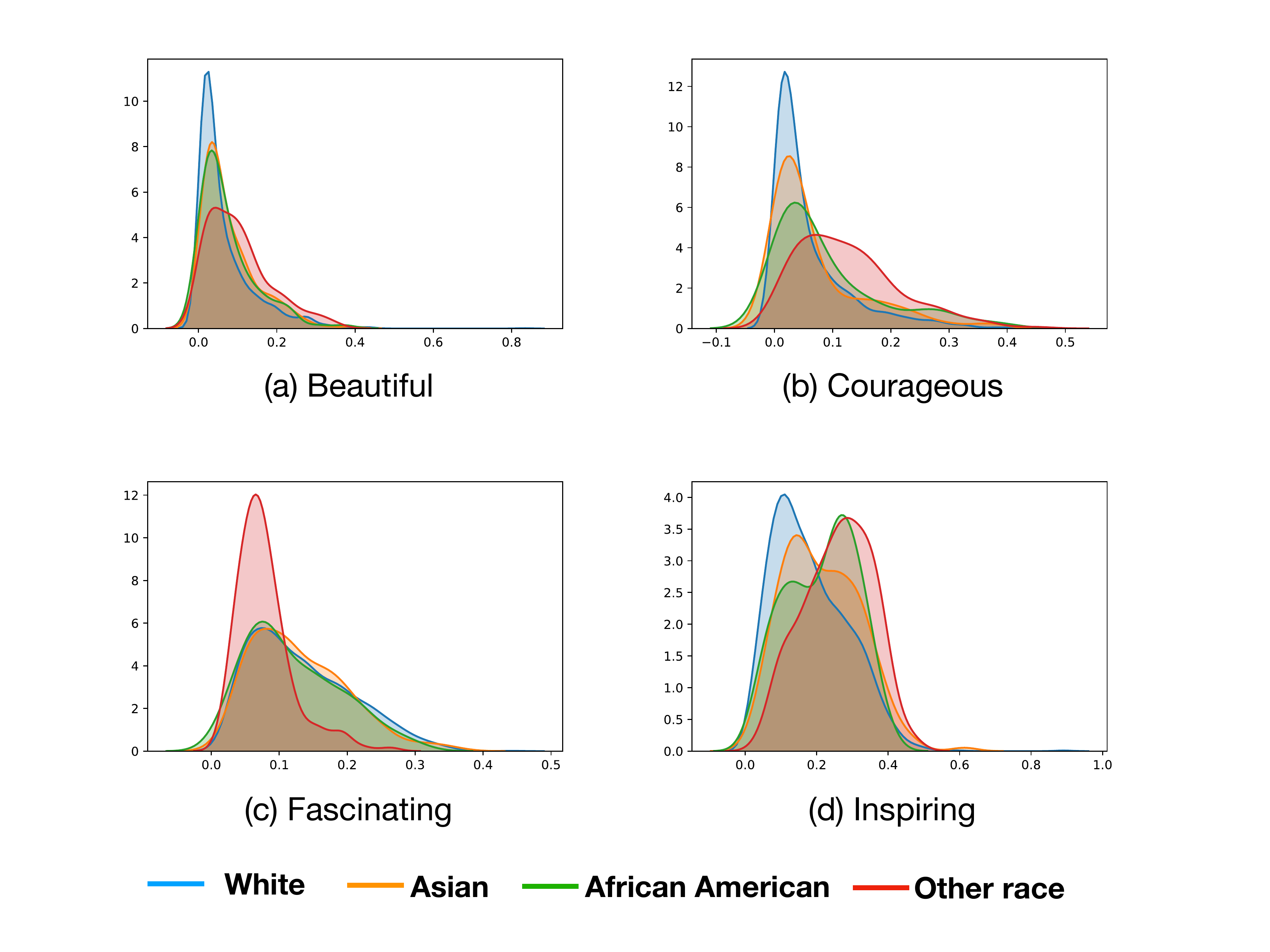} 
\caption{Comparison of the distributions of rating across different races. For different races, we have detected that some of the rating categories(here a) Beautiful, b) Couragous, c) Fascinating, d) Inspiring) have  different shape for the smoothed histogram. For example, talks of white speakers are rated to be beautiful and courageous with high confidence as compared to speakers of other races.}
\label{fig:race_smooth_hist}
\end{figure*}

\begin{figure*}
\centering
\includegraphics[width=1.\linewidth]{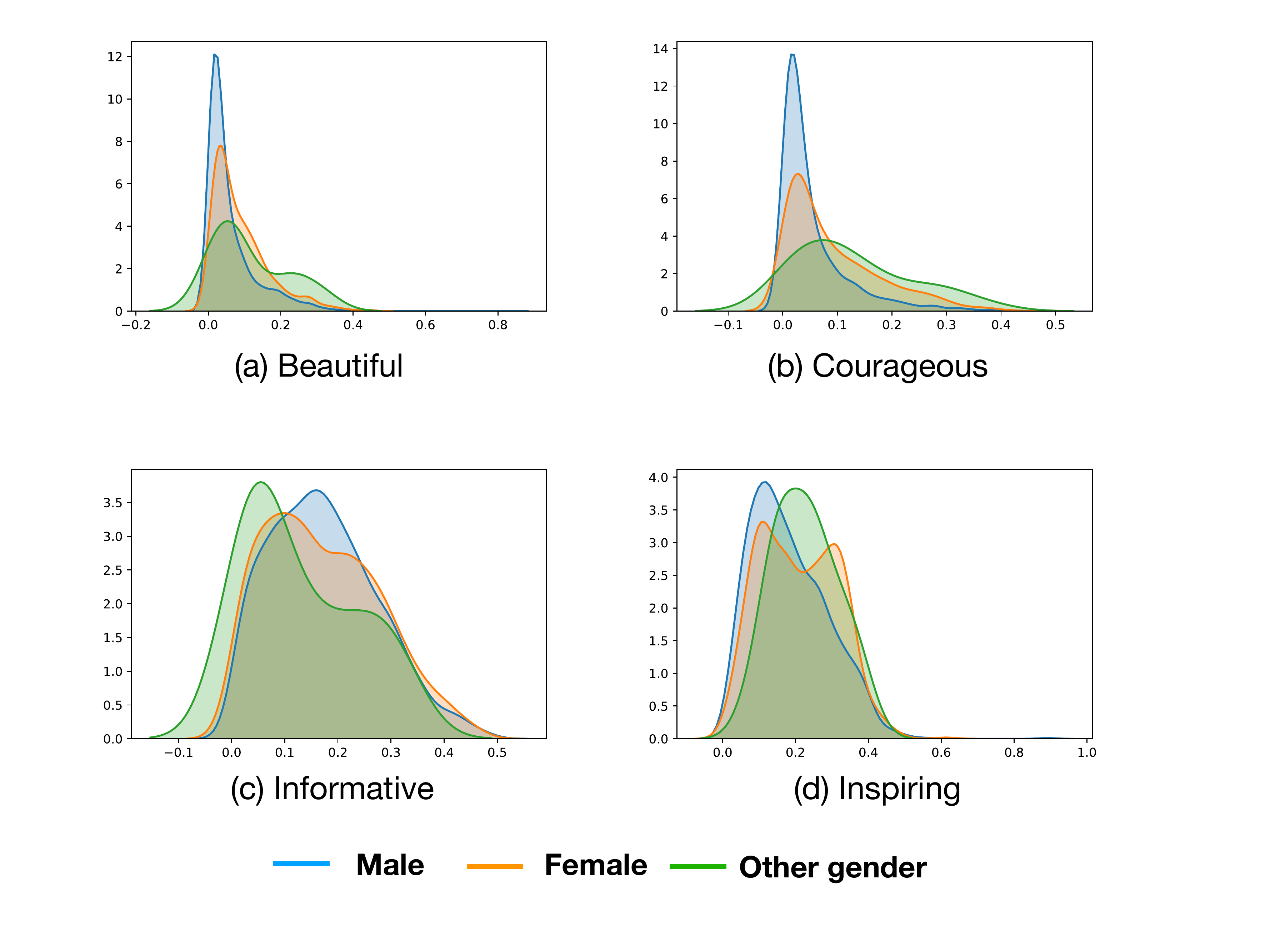} 
\caption{Comparison of distributions of rating across different genders. Similar to `race' attribute, here we also detected visible contrast among several rating categories across different genders. male speakers are confidently rated to have given beautiful and courageous talks when compared to any other gender}
\label{fig:gender_smooth_hist}
\end{figure*}

There is also a clear and visible difference in the distribution of the ratings for different groups corresponding to the protected attributes: race and gender. Figure~\ref{fig:race_smooth_hist} and figure~\ref{fig:gender_smooth_hist} represents smoothed histogram across different groups for several ratings where we observed substantial difference in the distribution. Interestingly we find one expected bias and one counter intuitive bias when considering ratings based on race. The expected bias is that talks of white speakers are rated to be beautiful and courageous with high confidence as compared to speakers of other races, the sharp blue distributions in top row of Figure \ref{fig:race_smooth_hist} are indicative of that. The wider green and red distributions on the top row of Figure \ref{fig:race_smooth_hist} confirm that the society is more confused and less confident while rating the talk of a non-white speaker as beautiful or courageous. On the other hand, counter intuitively, we find that viewers rate talks of speakers of other race as more fascinating than other races (sharpness of red distribution in \ref{fig:race_smooth_hist} in bottom row, left panel). Though counter intuitive, such bias is also not acceptable and clearly not fair. This highlights the diversity and difficulty of the issues of fairness and bias, since bias may not always stand out to be against the "expected" unprivileged class. Our work hence highlights the need to be careful when accounting for fairness as in fair society all types of biases should be removed. \\
Furthermore we also looked at the way viewers rate speakers based on gender. Even under this category we observed that male speakers are confidently rated to have given beautiful and courageous talks when compared to any other gender (sharpness, and less width of blue distributions in figure \ref{fig:gender_smooth_hist} indicates that). Though the confidence and tendency of rating in favor of male speakers drops substantially under informative and inspiring category (comparable width of green blue and red distributions) but is still slightly more than that for the speakers that are female (bi-modality of the distribution is an indication of confused rating behaviour) and of other gender. Besides highlighting the prevalence of biases with respect to gender and race, these results point out the need consider both expected and unexpected biases in data.\\
\begin{figure*}
\centering
\includegraphics[width=1.\linewidth]{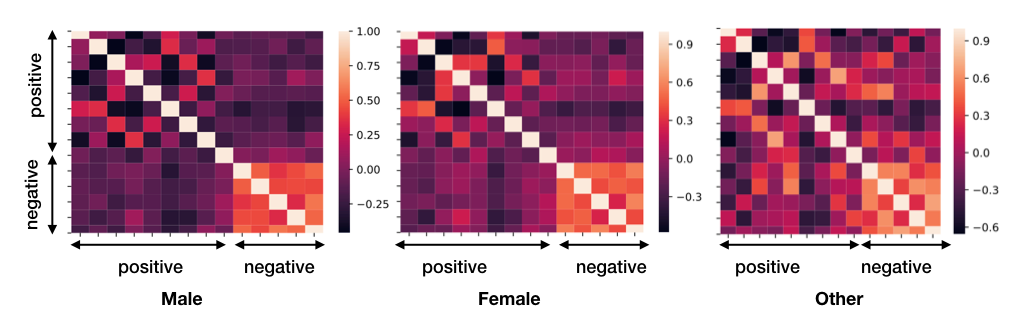} 
\caption{Correlation matrix heatmap across different genders. It is noticable that the male and female speaker's rating shows similar pattern in the correlation heatmap. On the other hand, the speakers from other gender shows a lot more variability in the heatmap. This can be identified as an indicator for the viewer's double minded intention when they rate 'other' gender's talks.}
\label{fig:correlation}
\end{figure*}
To dive deep into the nature of prevalent bias with respect to gender of a speaker we further divided the rating labels into positive and negative, where positive includes say, ``beautiful", ``courageous" etc. and negative includes say, ``confusing", ``unconvincing" etc. Now if we carefully notice \ref{fig:correlation}, we will see that there is lot more structure in ratings for male and female speakers and a lot more variability in the matrix for other speakers when positive and negative rating labels are considered. Note that, in our dataset a viewer can choose to rate with any of three possible labels. The chaos in the third matrix (rightmost) of figure \ref{fig:correlation} indicates that, for speakers of other genders, viewers tend to choose a mix of positive and negative labels. However for male and female speakers viewers are more sure and choose either all positive or all negative labels depending on their liking of the talk. This indicates that in general there is extreme confusion among viewers about their decision of whether or not they like a talk given by a speaker of other gender. This can also be identified as an indicator for the viewer's double minded intention when they rate those talks. 

\section{Methodology}

\begin{figure*}
\centering
\includegraphics[width=1.\linewidth]{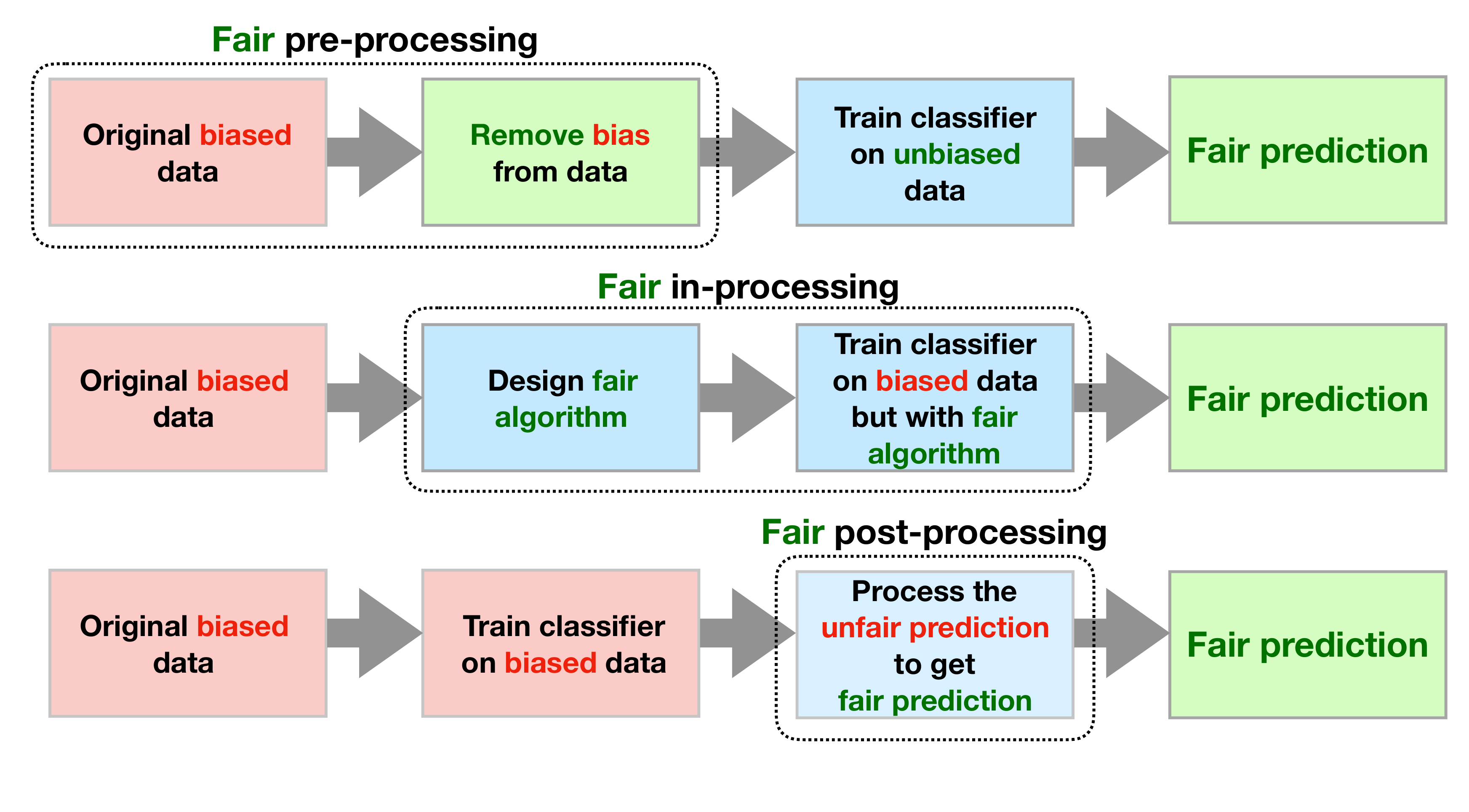} 
\caption{Standard pipeline for ensuring fairness in a dataset and machine learning models (adopted from). Fair pre-processing steps attempt to modify the data to detect bias before training any standard machine learning models. During in-processing steps, machine learning algorithm is modified (in most cases adding unfairness regularizer to the loss function) to learn a fair classifier. Post-processing steps include relabeling the predicted output to remove bias from the machine learning model.}
\label{fig:pipeline}
\end{figure*}

In this section we will describe the methods we used to design a bias free rating prediction. We explored all three methods as suggested in ~\cite{d2017conscientious}, 1) Pre-Processing, 2) In-Processing and 3) Post Processing.
We utilized the aforementioned toolkit AIF360 to implement these methods. This is explained in Figure \ref{fig:pipeline} which is adopted from~\cite{aif360-oct-2018}.
For each of the three steps, we quantify the amount of fairness with respect to a well defined metric, in this case \textit{`Disparate Impact'} ~\cite{biddle2006adverse}. In our TedTalk dataset, the fourteen rating categories are transformed to binary labels as described in Data section, i.e the category 'beautiful' can be 0 or 1 depending on if it was rated true or not. Using these binary labels we calculate the disparate impact in the rating of a video. Disparate impact can be understood with an example, let's say we consider the rating category `beautiful'. For female speakers, we can estimate how likely are they to get a beautiful rating. Disaprate impact compares such probabilities for all possible groups like male with female, female with other gender, male with other gender and so on. Disparate impact equals to 1 identifies as a fair case since it means both groups that are compared are equally likely to be rated beautiful.

\subsection{Pre Processing} As a first step we modify the original data before training the rating classifier. We then feed the modified and bias free data to the classifier for training as in \ref{fig:pipeline} top row. The main driving force for this step is the fact that raw data is inherently biased in almost all problems pertaining to social science. The pre-processing step attempts to make the data bias free, so that the classifier trained with this unbiased data makes fair predictions. When we look into our data in details, we see from Table \ref{tab:protected_attrinutes} that the there is a strong imbalance of bias in the data with respect to race and gender. We observe that our data has a stronger bias related to race than for gender. So with the goal to make fair predictions, we first preprocess the data and attempt to remove the dominant bias with respect to race.
\\
There are several ways to remove bias from the training data before feeding it to the classifier ~\cite{calmon2017optimized,feldman2015certifying,kamiran2012data,zemel2013learning}. 
We have chosen the disparate impact remover method ~\cite{feldman2015certifying} as the metric we have used is disparate impact metric.  We use the disparate impact remover from AIF360 for this purpose.  
\begin{figure*}
\centering
\includegraphics[width=0.7\linewidth]{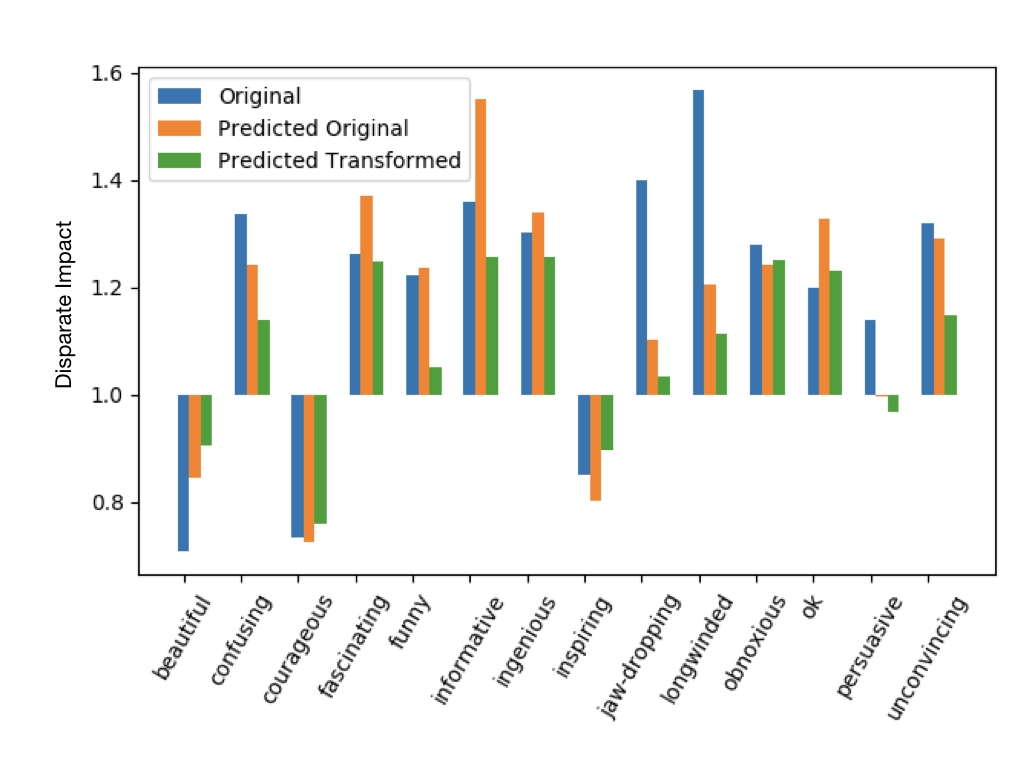} 
\caption{Disparate Impact metric between white and rest of the races improves after pre-processing. Values close to 1 represents fair classification for each rating category. The blue, green and red bars represents disparate impact of the original rating, predicted rating of classifier trained with original data and the modified data after pre-processing respectively. Observe that the disparate impact is improving for most the rating labels. }
\label{fig:di_remover}
\end{figure*}
The result of the preprocesing step is shown in Figure~\ref{fig:di_remover}. This figure is showing the disparate impact across all rating labels of the dataset. We computed the disparate impact metric for the original label of the dataset and observed that it is far from 1 for most of the rating categories as shown by the blue bars. This exhibits the amount of unfairness present in the original dataset. The huge existing bias in original data highlights the need to design a fair predictor of the rating. Hence, we trained a logistic regression model with both the original dataset and the pre-processed dataset. We observed that prediction on preprocessed data has disparate impact closer to 1 (Figure \ref{fig:di_remover}) than when it is trained with the orginal biased data. Our result hence shows that it is important to identify the cause of dominant bias in the data and remove it by preprocessing to gain substantial improvement in fair predictions.

\subsection{In Processing}
\begin{figure*}
\centering
\includegraphics[width=0.7\linewidth]{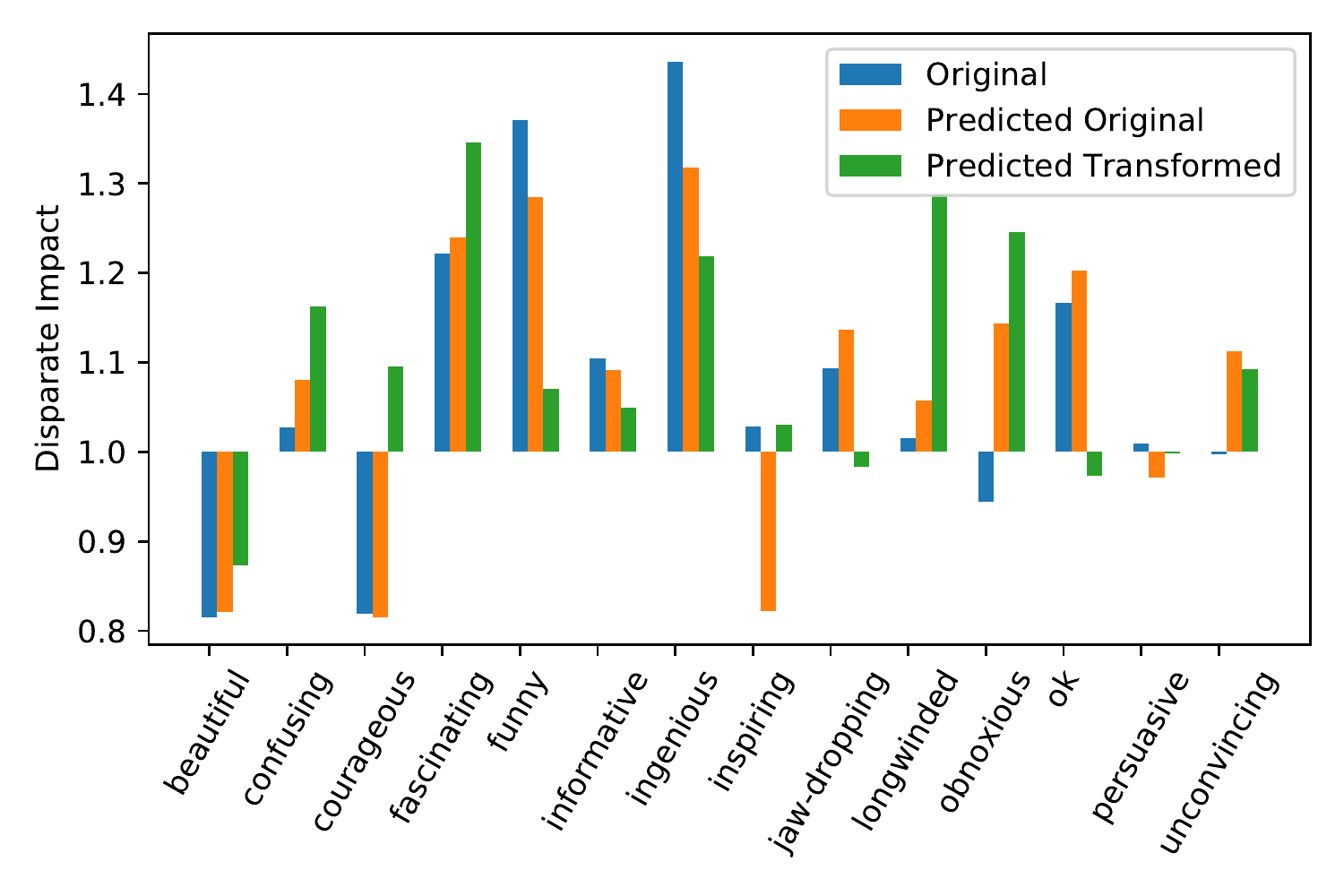} 
\caption{Disparate Impact metric between male and rest of the gender improves after in-processing. Observe that the disparate impact is improving for most the rating labels.}
\label{fig:inproc}
\end{figure*}

In the last section we showed that it is possible to build a fair model by removing the bias present in the original data before training a classification model. However it is not always possible to do preprocessing, because there may not be access to the original data or it might be hard to identify the cause of bias in the data or it may be time consuming to re-annotate data and so on. Under such scenarios the solution is to design a fair classifier that uses a fair algorithm instead. So this allows us to still train the model with a biased dataset as input but the predictions employed by the fair algorithm are fair. The algorithm and the classifier in this case identifies the presence of bias in the dataset used for training and adjusts appropriately (based on the amount of bias) while predicting corresponding labels. Examples of such in processing techniques for bias mitigation can be found in ~\cite{art2018,kamishima2012fairness,zhang2018mitigating}. For example, in our case, we hope that even if there is some bias in our dataset in favor of male, our rating classifier will make sure it accounts for that and weighs females and speakers of other genders equally to get rid of such unwanted unfairness. This will then generate fair rating predictions.\\
For the TED talk data we have used the ``Prejudice Remover" to do in processing ~\cite{kamishima2012fairness}. The intuition is that we attempt to make the dependence of rating on sensitive attributes as strong as the nonsensitive attributes. For example, assume that female speaker is unprivileged and male speaker is privileged, that is ratings are heavily biased by male speakers in the real data. In that case the goal of the classifier and the learning algorithm is to make sure that rating predictions depend equally on male and female speaker and is not strongly influenced by male speakers only instead even if that is the case in the biased data.\\
The result of in-processing is shown in Figure~\ref{fig:inproc}. This figure is showing the disparate impact across all rating labels of the dataset. Similar to the pre-processing part we computed the disparate impact metric (in this case we compute the metric for gender bias) for the original label of the dataset and observed that it is far from 1 for most of the rating categories as shown by the blue bars,. This exhibits the amount of unfairness present in the original dataset w.r.t gender. We then trained a logistic regression model and a Prejudice Remover model \cite{kamishima2012fairness} with the original dataset. We observed that prediction of the Prejudice Remover model has disparate impact closer to 1 (Figure \ref{fig:di_remover}) than the prediction of the logistic regression model for most of the categories.

\subsection{Post Processing}
We tried many popular methods ~\cite{hardt2016equality,pleiss2017fairness}  available in literature to employ post processing in our data. However, none of the methods gained significant improvement in making fair predictions. This highlights two important issues: 1) Nature of bias strongly drives what technique works best in terms of making fair predictions, in our case gender and race bias cannot be removed by just post processing technique 2) Not all types of dataset can used to make fair prediction with a fixed processing technique, in our case, as shown, post processing is unsuitable for the TED talk data videos.

\subsection{Conclusion from analysis}
We have employed a principled technique for removing bias and making fair rating predictions for TED talk videos. We first identify the dominant cause of the bias (in our case it is racial bias) in the data and remove it by pre processing, we then remove the other non-dominant causes of bias (gender bias in our case) by in-processing and we show that post processing does not make any improvement in fair predictions for our data. This highlights that choice of processing to achieve fair predictions heavily depends of types of bias and the dataset being used. This establishes the need to explore all possible biases and all possible techniques to obtain best results in terms of fairness. 

\section{Discussion}
Ensuring fairness in a machine learning model is a complex, multi-faceted issue which depends not only on the challenges presented by the immediate task but also on the definition of fairness that is in use. Unfairness or bias may arise at any stage of a machine learning pipeline. Thus, while developing machine learning algorithms which can make fair predictions for a classification task is important, creating a data collection process that is free from bias is essential for the model's ultimate success. For example, Holstein et al.~\cite{holstein2019improving} show that commercial machine-learning product teams express the need for tools that can actively guide the data collection process to ensure fairness in the downstream machine learning task. In our study, we demonstrate a systematic method of detecting bias using the AIF360 toolkit and show that various bias mitigation techniques can be applied to our data set at various stages of a machine learning pipeline (i.e. pre-processing, in-processing, post-processing). This procedure can be followed by any machine learning practitioner or data scientist who has to deal with real world data to make machine learning models for decision making.\\
We have applied a state of the art fairness measuring metric `Disparate impact' to a new and diverse dataset `TedTalk', revealing sharp differences in the perception of speakers based on their race and gender. Based on our findings, we demonstrate that in TED Talk ratings, viewers rate `white male' speakers confidently, while all other speakers are rated in a weaker manner. On the other hand, viewers rating shows a great deal variability when rating any other groups combined. \\
Our study shows that detecting and erasing bias in the data collection process is essential to resolving issues related to fairness in AI. While the existing literature about fairness in AI~\cite{dwork2012fairness,pedreschi2009measuring,russell2017worlds} have focused almost exclusively on detecting bias in specific domains such as recidivism prediction, admission decision-making, face detection etc, the analysis of implicit bias and . In all these scenarios, researchers come up with a quantifiable measurement of fairness and train the model which ensures the model is fair. However, in a more diverse domain where the users and system interacts in a complex way (e.g. recommendation system, chatbots etc), there is a lot of work left to do to detect and erase unfairness.

While we have evaluated the Ted Talk dataset with specific metrics that are appropriate to it, there is an urgent need to develop  metrics that are adapted just as uniquely to the needs of each dataset and its possible implementations. Identifying different kinds of social bias that affects in applications directly engaging with social influence and opinion formation is absolutely necessary for creating fair and balanced artificial intelligence systems.

\newpage
\bibliographystyle{SIGCHI-Reference-Format}
\bibliography{reference}

\end{document}